\documentclass[journal]{IEEEtran}
\usepackage{bm}
\usepackage{cite}
\usepackage{algorithm}  
\usepackage{algorithmic}
\usepackage{graphicx}
\usepackage{booktabs}
\usepackage{diagbox}
\usepackage{amssymb}
\usepackage{times}
\usepackage{epsfig}
\usepackage{amsmath}
\usepackage{algorithm}  
\usepackage{algorithmic} 
\usepackage{enumerate}
\usepackage{multirow}
\usepackage{verbatim}
\ifCLASSINFOpdf
\else
\fi
\begin{document}
\title{Preparing Lessons: Improve Knowledge Distillation with Better Supervision}

\author{Tiancheng Wen,~\IEEEmembership{}
        Shenqi Lai,~\IEEEmembership{}
        Xueming Qian*~\IEEEmembership{}
        % <-this % stops a space
\thanks{Authors Tiancheng Wen and Shenqi Lai contribute to this work.}% <-this % stops a space
\thanks{This work was supported in part by the NSFC under Grant 61732008, 61772407, the World-Class Universities(Disciplines) and the Character-istic Development Guidance Funds for the Central Universities, under No.PY3A022.}% <-this % stops a space
\thanks{Tiancheng Wen (E-mail: ssstormix@stu.xjtu.edu.cn) and Shenqi Lai (E-mail: laishenqi@stu.xjtu.edu.cn) are with the SMILES LAB at School of Electronics and Information Engineering, Xi’an Jiaotong University, Xi’an, Shaanxi 710049, China. Xueming Qian (corresponding author, qianxm@mail.xjtu.edu.cn) is with the Ministry of Education Key Laboratory for Intelligent Networks and Network Security and with SMILES LAB, Xi’an Jiaotong University, 710049, Xi’an, China.}}

\maketitle

\begin{abstract}
Knowledge distillation (KD) is widely used for training a compact model with the supervision of another large model, which could effectively improve the performance. Previous methods mainly focus on two aspects: 1) training the student to mimic representation space of the teacher; 2) training the model progressively or adding extra module like discriminator. Knowledge from teacher is useful, but it is still not exactly right compared with ground truth. Besides, overly uncertain supervision also influences the result. We introduce two novel approaches, Knowledge Adjustment (KA) and Dynamic Temperature Distillation (DTD), to penalize bad supervision and improve student model. Experiments on CIFAR-100, CINIC-10 and Tiny ImageNet show that our methods get encouraging performance compared with state-of-the-art methods. When combined with other KD-based methods, the performance will be further improved.
\end{abstract}


% \fi
\IEEEpeerreviewmaketitle

\section{Introduction}
\IEEEPARstart{K}{nowledge} Distillation (KD) methods have drawn great attention recently, which are proposed to solve the contradiction between accuracy and depolyment. The teacher-student techniques use "knowledge" to represent the recognition ability of deep model. Thus, adopting teacher's knowledge as supervision will guide student to have more discrimination. To improve transfer efficiency, many recent related papers focus on designing different kinds of knowledge \cite{ahn2019variational,chen2018learning,heo2019comprehensive,heo2019knowledge,huang2017like,park2019relational,romero2014fitnets,tung2019similarity,wang2018progressive,yim2017gift,zagoruyko2016paying,guo2019robust,passalis2018unsupervised}, or extending training strategies \cite{chen2018knowledge,furlanello2018born,gao2018embarrassingly,mirzadeh2019improved,shen2019meal,wang2018progressive,xu2019data,xu2017training,yang2018knowledge,yuan2019revisit,zhang2019your,zhao2019highlight,yuan2019ckd,passalis2018unsupervised}. The works have obtained positive results.

Though teacher usually has stronger discrimination than student, there still exist incorrect or uncertain predictions. Using these kinds of knowledge as supervision will lead to bad result for student model. Most previous works hold little discussion about the topic. Adding a cross entropy loss with ground truth or using teacher ensemble can alleviate this problem. But both of them do not handle it from the source and still cannot guarantee validity of supervision.

On the other hand, some works \cite{szegedy2016rethinking, xie2016disturblabel, zhang2017mixup} indicate that uncertainty of supervision may be beneficial to network training. But in bad supervision case of KD, uncertainty concentrates at the same training samples in each iteration, while the benefitial uncertainty should act randomly.

In this paper, we revisit KD framework and analysis the phenomenon of bad supervision. We propose two simple and universal methods, Knowledge Adjustment (KA) and Dynamic Temperature Distillation (DTD), to obtain better supervision. Fig. 1 describes our idea, teacher could make incorrect predictions and uncertain predictions towards some samples, and the proposed KA and DTD methods are respectively targeted to the two problems. For KA, the predictions of misjudged training samples are fixed according to corresponding ground truth before loss computation. Student model trained in this way is fed completely correct information from teacher model. For DTD, we find that some uncertainty of soft targets comes from excessive temperature, thus we turn to design dynamic sample-wise temperature to compute soft targets. By conducting so, student training will receive more discriminative information on confusing samples. From this perspective, DTD can be viewed as a process of (online) hard example mining.

\begin{figure}[t]
	\begin{center}
		\includegraphics[width=8cm]{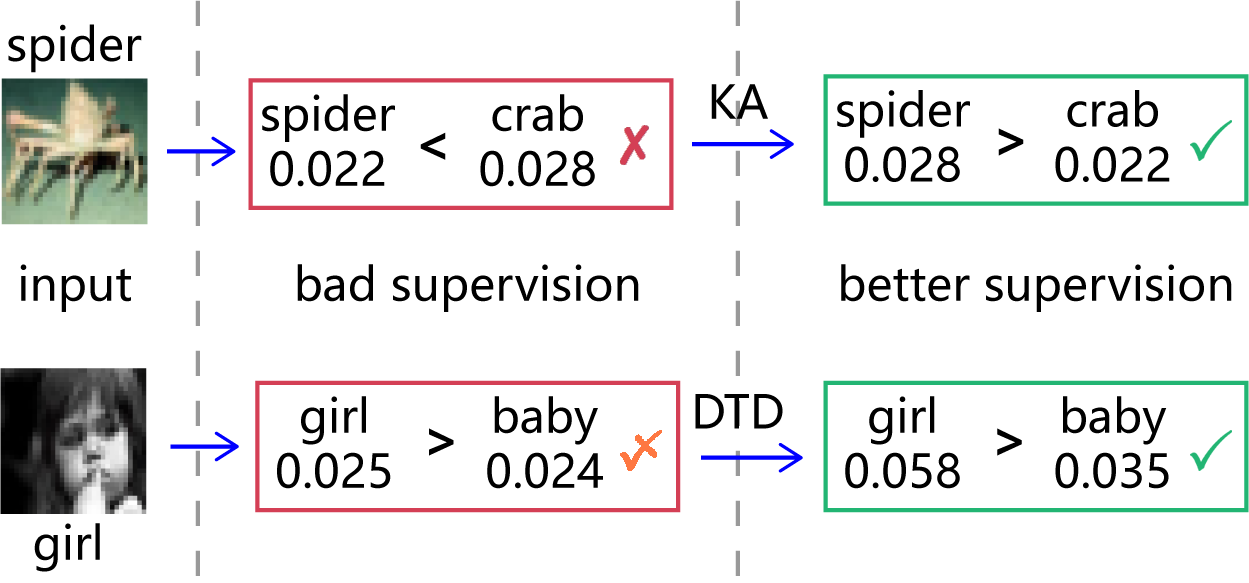}
	\end{center}
	\caption{Illustration of the proposed methods. The numbers denote the teacher predicting probabilities of corresponding classes. KA deals with incorrect supervision and DTD handles uncertain supervision.}
	\label{fig:fig1}
\end{figure}

Our contributions of this work are summarized as follows:

{\bf 1.} We give analysis about the bad supervision problem of existing KD-based methods, including teacher's incorrect predictions and overly uncertain predictions. We find that nearly half the errors of student model are related to those of teacher model, and manually given temperature parameter may lead to uncertain supervision.

{\bf 2.} Knowledge Adjustment is proposed to handle wrong predictions from teacher model. Two distinct implementations are used. Both of them will make the incorrect predictions go away and generate completely correct supervision for student.

{\bf 3.} We propose Dynamic Temperature Distillation to avoid overly uncertain supervision from teacher model. We also give two implementations to achieve the goal. More certain but still soft supervision will make student model more discriminative.

{\bf 4.} We evaluate the proposed methods on CIFAR-100, CINIC-10 and Tiny ImageNet. The experimental results show they can not only improve performance independently, but also get better results when combined together. Furthermore, when we adopt KA, DTD and other KD-based methods together, the final result could even achieve the state-of-the-art.

\section{Related Works}
{\bf Knowledge Distillation and Transfer.} Strategy of transferring knowledge from one neural network to another has developed for over a decade. It has been proved that a shallow neural network can achieve better performance with a well-trained cumbersome network as auxiliary. As far as we know, Bucilua \textit{et al.} \cite{bucilua2006model} firstly explored training a single neural network to mimic an ensemble of networks. Ba and Caruana \cite{ba2014deep} further compress complex networks by adopting logits from cubersome model instead of ground truth as supervision. Inspired by \cite{ba2014deep}, Hinton \textit{et al.} \cite{hinton2015distilling} propose the method of KD. With keeping the basic teacher-student framework, KD modifies the softmax function with an extra parameter $\tau$. Then logits from both teacher model and student model are softened. Student is trained to mimic the distribution of teacher soft targets.

Based on the above works, many recent studies focus on designing different knowledge representation for efficient transfer. Romero \textit{et al.} \cite{romero2014fitnets} use "hints" to transfer knowledge, which are actually feature maps produced by some chosen teacher's hidden layers. Yim \textit{et al.} \cite{yim2017gift} transfer the variation of feature maps from both ends of some sequential blocks. In \cite{zagoruyko2016paying}, activation-based attention and gradient-based attention are investigated to carry out knowledge transfer. Huang and Wang \cite{huang2017like} match the neuron activation distribution to carry out distillation. Park \textit{et al.} \cite{park2019relational} explore sample relations as knowledge carrier. Tung and Mori \cite{tung2019similarity} point out that inner-batch similarities help to train student better. Chen \textit{et al.} \cite{chen2018learning} realize transfer with feature embedding. Similar to \cite{tung2019similarity} and \cite{chen2018learning}, Passalis and Tefas \cite{passalis2018unsupervised} make student network learn the similarity embeddings. Ahn \textit{et al.} \cite{ahn2019variational} propose to maximize mutual information between teacher and student.  And Guo \textit{et al.} \cite{guo2019robust} introduce an objective which minimizes the difference between gradients between student and teacher model.

There are also works improving or extending KD with training strategies. Furlanello \textit{et al.} \cite{furlanello2018born} , Mirzadeh \textit{et al.} \cite{mirzadeh2019improved} and Gao \textit{et al.} \cite{gao2018embarrassingly} propose multi-stage training scheme. Xu \textit{et al.} \cite{xu2017training}, Chen \textit{et al.} \cite{chen2019data} and Shen \textit{et al.} \cite{shen2019meal} introduce adversarial learning into KD framework, adopting GAN or part of GAN as an auxiliary. Self-distillation is discussed in \cite{xu2019data}, \cite{yuan2019revisit} and \cite{zhang2019your}, in which knowledge is exploited inside a single network. Moreover, Yuan \textit{et al.} \cite{yuan2019ckd} propose collaborative distillation, training two models with mutual knowledge rather than teacher-student framework.

KD has been envolved in many specific tasks. For instance, Ge \textit{et al.} investigate recognition with the help of KD, in which work a teacher recognising high-resolution face guides student to recognise low-resolution face. Han \textit{et al.} \cite{han2019neural} use teacher model to encode knowledge of fashion domain, then student is guided to perform on clothes matching. Li \textit{et al.} \cite{li2019spatiotemporal} deploy KD in video saliency task, transferring spatial-temporal knowledge to simple student netowrks.

{\bf Label Regularization.} Regularization is common practice to deal with over-fitting in network training. Label Regularization focuses on supervision fed to network, which designs more complicated labels instead of traditional one-hot label, encouraging deep models to generalize and exploit inter-class information. Szegedy \textit{et al.} \cite{szegedy2016rethinking} propose Label Smoothing Regularization (LSR) to give the non-ground-truth labels tiny values. Xie \textit{et al.} \cite{xie2016disturblabel} add noise to label. They random replace some labels with incorrect ones to prevent overfitting. Recently, updating labels iteratively has been widely investigated. Bagherinezhad \textit{et al.} \cite{bagherinezhad2018label} train Label Refinery Network to modify labels at every epoch of training. Ding \textit{et al.} \cite{ding2019adaptive} use dynamic residual labels to supervise the network training. Moreover, label regularization and KD may share many characteristics. Yuan \textit{et al.} \cite{yuan2019revisit} investigate the relation between label regularization and KD. The work interprets KD as a learned label regularization term rather than manually designed one. 

{\bf Hard Example Mining for Deep Models.} Hard example mining has brought discussions recently, which treat training samples unbalancedly, paying more attention to those imperfectly learned samples. Shrivastava \textit{et al.} \cite{shrivastava2016training} propose Online Hard Example Mining (OHEM), selecting samples according to their losses and refeeding network those samples. Wang \textit{et al.} \cite{wang2017fast} generate hard examples via GAN instead of filtering the original data. Method of Focal Loss \cite{lin2017focal} weights samples with gamma nonlinearity about their classification scores. And Li \textit{et al.} \cite{li2019gradient} investigate hard examples' gradients and design Gradient Harmonizing Mechanism (GHM) to solve the imbalance in difficulty.

\begin{figure*}[t]
	%\begin{center}
	\includegraphics[width=17cm]{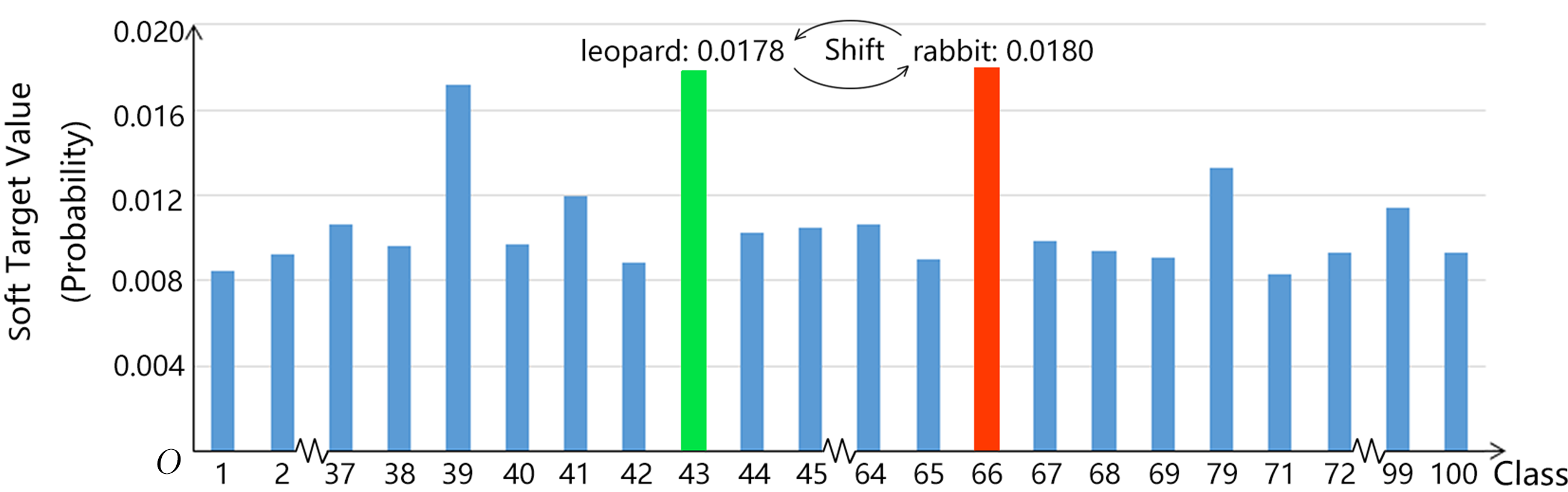}
	%\end{center}
	\caption{PS on a misjudged sample's soft target. The image is from CIFAR-100 training data, whose ground truth label is "leopard" but ResNet-50 teacher's prediction is "rabbit". The value of "leopard" is still large, which indicates that the teacher does not go ridiculous. Shift operation is carried out towards these two classes.}
\end{figure*}

\section{The proposed method}

\begin{figure*}[t]
	%\begin{center}
	\includegraphics[width=17cm]{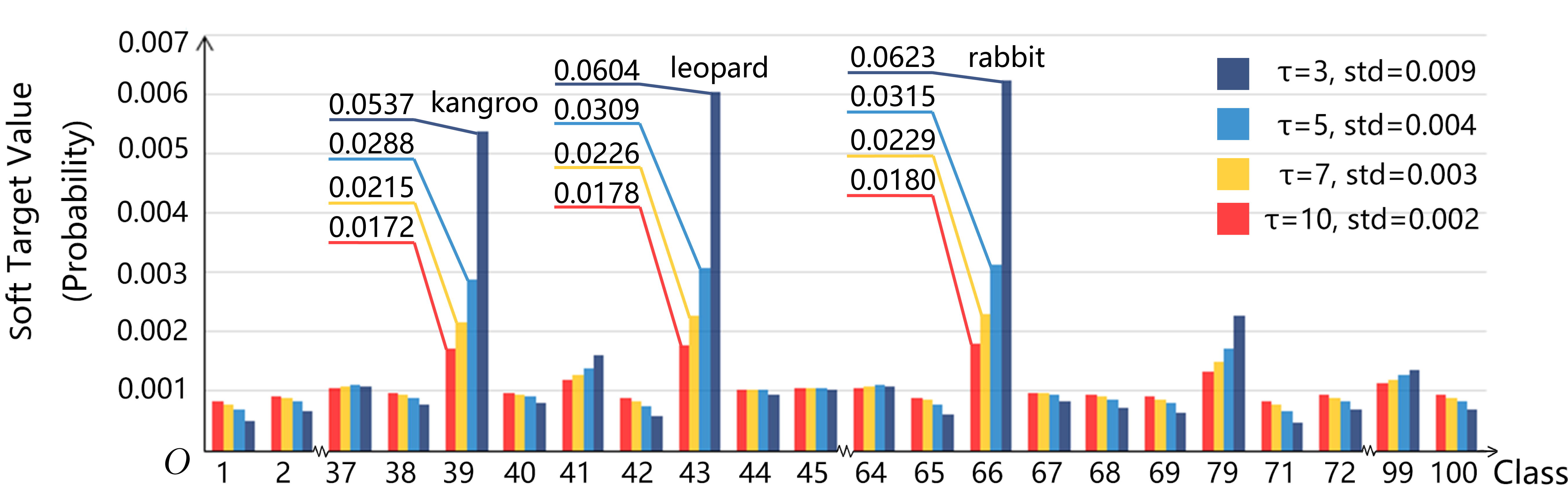}
	%\end{center}
	\caption{Soft target's distributions with different $\tau$. The image is from CIFAR-100 training set, whose ground truth label is "leopard". Inter-class difference gets small when $\tau$ goes up. Specifically, values of "kangaroo", "leopard" and "rabbit" get extremely close at high temperature, which makes "kangaroo" and "rabbit" disturbance items during training. Thus a relatively lower $\tau$ is proper.}
\end{figure*}

In this section, we firstly investigate two bad phenomenons existing in KD and KD-based methods, then introduce corresponding solutions named as Knowledge Adjustment (KA) and Dynamic Temperature Distillation (DTD). For both of them, we propose two distinct implementations.

\subsection{Genetic errors and knowledge adjustment}
In KD and KD-based methods, the student model is trained under the predictions of teacher model, no matter if it is true or not. We use "Genetic Error" to note student's incorrect predictions that are same with teacher's. Obviously when teacher makes mistakes, student can hardly correct the error knowledge itself, thus genetic errors occur.

First we take a review of soft targets from KD. For a sample $x$  with ground truth $\bm{y}$, KD uses softmax function with temperature parameter $\tau$ to soften the student's logits $\bm{v}$ and teacher's logits $\bm{t}$, the $\bm{i}$-th output is given by Eq. (1)
\begin{equation}\label{(1)}
\bm{p}_{\tau, i}=\frac{exp(\bm{v}_i/\tau)}{\sum_{j}exp(\bm{v}_j/\tau)}. 
\end{equation}

We denote KD loss using cross entropy between student's $\tau$ softened logits $\bm{p}_\tau$ and teacher's $\tau$ softened logits $\bm{q}_\tau$.
\begin{equation}\label{(2)}
\mathcal{L}_{KD}=\alpha\tau^2\mathcal{H}(\bm{q}_\tau, \bm{p}_\tau)+(1-\alpha)\mathcal{H}(\bm{y}, \bm{p}_\tau), 
\end{equation}
where $\mathcal{H}(\cdot)$ denotes cross entropy.

The pretrained teacher model can not guarantee $\bm{t}$ is always right, whether the supervision exists in soft targets or feature maps. Actually Eq. (2) and some other methods \cite{heo2019comprehensive,huang2017like,tung2019similarity,wang2018progressive} add an extra cross entropy with ground truth to alleviate this situation. However, this kind of correction is slight and there remains a noteworthy proportion of genetic errors. For a ResNet-18 student educated (KD) by a ResNet-50 teacher on CIFAR-100, the genetic errors make up 42.4\% of total 2413 misjudged samples in the test set.

We then propose KA, trying to fix the prediction of teacher with reference to the ground truth label. The exact operation is conducting an adjusting function $A(\cdot)$ on teacher's softened logits, and the modified loss is expressed with KL divergence
\begin{equation}\label{(3)}
\mathcal{L}_{KA}=\tau^2KL(A(\bm{q}_\tau), \bm{p}_\tau). 
\end{equation}

$A(\cdot)$ has following three charactics. Firstly, $A(\cdot)$ will fix the incorrect $\bm{q}_\tau$ and do nothing on correct ones. And a better $A(\cdot)$ tends to make slight modification on incorrect predictions. Secondly, It receives a single parameter $\bm{q}_\tau$, which indicates $A(\bm{q}_\tau)$ is invariable during student training, so the optimization object is equivalent to cross entropy. Thirdly, we don't need the cross entropy computed with $\bm{y}$, for $A(\bm{q}_\tau)$ is totally correct. Based on the principles above, we adopt Label Smooth Regularization (LSR) \cite{szegedy2016rethinking} and propose Probability Shift (PS) to implement $A(\cdot)$.

{\bf Label Smooth Regularization.} At the start, LSR is proporsed as model regularization method to overcome over-fitting \cite{szegedy2016rethinking}. Considering a sample $x$ of class $k$ with ground truth label distribution $l=\delta(k)$, where $\delta(\cdot)$ is impulse signal, the LSR label is given as
\begin{equation}\label{(4)}
l^\prime=(1-\epsilon)\delta(k)+\epsilon/K, 
\end{equation}

$K$ is number of classes. Compared with one-hot label distribution, It gives a less confident probability (but still the most confident among all the classes) to the ground truth label, and allocates the remainder of the probability equally to other classes. The mechanism is similar to KD. And from this point of view, LSR actually provides a simple technique to soften one-hot label, which can be used to realize $A(\cdot)$. In KA, we directly replace the incorrect soft targets with Eq. (4). Meanwhile the correct soft targets stay the same. In addition, LSR replacements should get involved after distilling softmax to ensure logits to be softened only once.

{\bf Probability Shift.} Given an incorrect soft target, PS simply swaps the value of ground truth (the theoretical maximum) and the value of predicted class (the predicted maximum), to assure the maximum confidence is reached at ground truth label. The process is illustrated in Fig. 2. PS operation adjusts the theoretical maximum (leopard) and the predicted maximum (rabbit). It keeps all the probabilities produced by teacher model instead of discarding wrong ones. And the fixed tensor keeps the overall numerical distribution of the soft target.

Compared with directly replacing soft targets with ground truth, both LSR and PS retain some dark knowledge from tiny probability, which is pointed out to be useful according to \cite{hinton2015distilling}. The methods also keep the numerical distribution of soft targets, which is helpful to stabilize the training process.

In fact, the incorrect predicted class often shares some similar features with the sample. That is to say, the incorrect predicted class may contain more information than the other classes. In addition, teacher's misjudgments come from global optimization covering all the training samples, and inordinate partial corrections may break the convergence of student model. Therefore, PS is a more promising solution compared with LSR.
%\IEEEpubidadjcol

\subsection{Dynamic temperature distillation}
Although previous works \cite{ba2014deep,hinton2015distilling,sau2016deep} indicate that student can benefit from uncertainty of supervision, overly uncertain predictions of teacher may also affect the student. We analysis this problem from perspective of soft targets and temperature $\tau$. The effect of distilling softmax is visualized in Fig. 3. The distribution of soft logits becomes "flat" when $\tau$ is set more than 1. This indicates that the supervision may lose some discriminative information with high temperature. Therefore, student may be confused on some samples who get significant and similar scores on several classes. As a solution, we propose a method named as Dynamic Temperature Distillation (DTD), to customize supervision for each sample.

The basic idea of DTD is to make $\tau$ vary on training samples. Especially for samples who cause confusion easily, $\tau$ should be smaller to enlarge inter-class similarity. And for easily learned samples, as \cite{hinton2015distilling} points out, a bigger $\tau$ will help to utilize the classificatory information about the classes. The object function of DTD can be expressed as
\begin{equation}\label{(5)}
\mathcal{L}_{DTD}=\alpha{\tau_x}^2KL(\bm{q}_{\tau_x}, \bm{p}_{\tau_x})+(1-\alpha)\mathcal{H}(\bm{y}, \bm{p}_{\tau_x=1}),
\end{equation}
where $\tau_x$ is sample-wise temperature, and we use Eq. (6) to compute $\tau_x$.
\begin{equation}\label{(6)}
\tau_x=\tau_0+(\frac{\sum_j\bm{\omega}_j}{N}-\bm{\omega}_x)\beta,
\end{equation}
where $N$ is batch size, $\tau_0$ and $\beta$ denote the base temperature and bias. And $\omega_x$ is batch-wise normalized weight for sample $x$, describing the extent of confusion. $\omega_x$ is designed to increase	 when $x$ is confusing and teacher's prediction is uncertain. Then $\tau_x$ is computed to be smaller than $\tau_0$, and soft target gets more discriminative. In addition, note that Eq. (5) is written in KL divergence rather than cross entropy, since $\bm{p}_{\tau_x}$ varies with $\tau_x$ and no longer a constant.

As analyzed above, confusing samples should get larger weights and further get lower temperature, which leads to more discriminative soft targets. In this way, DTD selects confusing samples and pays more attention to them, which can also be viewed as hard example mining.

Then we propose two methods to obtain $\omega_x$. Inspired by Focal Loss \cite{lin2017focal}, we propose Focal Loss Style Weights (FLSW) to compute sample-wise weights. Our another method computes $\omega_x$ following the max output of student's prediction, which we call Confidence Weighted by Student Max (CWSM).

{\bf Focal Loss Style Weights.} The original focal loss is proposed towards the task of object detection and defined as Eq. (7)
\begin{equation}\label{(7)}
FL(p)=-(1-p)^\gamma log(p),
\end{equation}
where $p$ is the classification score of one sample, denoting the model confidence towards the ground truth category. $(1-p)^\gamma$ actually differentiates samples with learning difficulty. A hard sample makes more contribution to the total loss, so the model pays more attention on hard samples during training.

In out method, the learning difficulty can be measured with the similarity between student logits $\bm{v}$ and teacher logits $\bm{t}$. Then we can denote $\bm{\omega}_x$ as
\begin{equation}\label{(8)}
\bm{\omega}_x=(1-\bm{v}\cdot\bm{t})^\gamma.
\end{equation}

$\bm{v}\cdot\bm{t}\in[-1, 1]$ is the inner product of two distributions, which we can use to evaluate student's learning about the sample. $\bm{\omega}_x$ goes relatively big when student's prediction is far from teacher's, which meets the monotonicity discussed before.

{\bf Confidence Weighted by Student Max.} We also weight samples by the max predictions of student, which to some extent reflects the learning situation of samples. Confusing samples generally have uncertain predictions. And we assume it is teacher's undemanding supervision that leads to student's uncertain predictions. In CWSM, $\bm{\omega}_X$ is computed using Eq. (9)
\begin{equation}\label{(9)}
\bm{\omega}_x=\frac{1}{\bm{v}_{max}},
\end{equation}
and student logits $\bm{v}$ should be normalized. $\bm{v}_{max}$ can be deemed to represent student's confidence towards the sample. Obviously, samples with less confidence get larger weights. And gradients from these samples will contribute more during the distillation.

$\bm{\omega}_x$ computed following whether FLSW or CWSM should not be directly involved into computation of the overall loss due to some numerical issue. Specifically, $\bm{\omega}_x$ may get tiny for all the samples in one batch along training, because most of the training samples can be predicted correctly by deep model. And the loss in this batch is small. However, whether a sample is "easy" or "hard" is a relative value. So we conduct L1 normalization to make $\bm{\omega}_x$ numerically comparable and controllable.

\subsection{Total loss function and algorithm}

In this section, we combine KA and DTD together, the overall loss function is designed as Eq. (10).
\begin{equation}\label{(10)}
\mathcal{L}_{total}={\tau_x}^2KL(A(\bm{q}_{\tau_x}), \bm{p}_{\tau_x}),
\end{equation}

Eq. (10) is similar to Eq. (3), but the difference is that Eq. (10) uses sample-wise temperature to soften logits. It is worthy to point out that the supervision tensor $A(\bm{q}_{\tau_x})$ varies with the learning situation, so the KL divergence is not equivalent to cross entropy. In addition, the cross entropy with ground truth is needless here, because $A(\bm{q}_{\tau_x})$ is totally correct and have already utilized ground truth information.

$\bm{\omega}_x$ brings gradients when optimizing student network. In limited varying space, $\mathcal{L}_{total}$ goes low when temperature increases, because high temperature closes the KL divergence between soft logits from teacher and student. Thus minimization of $\mathcal{L}_{total}$ helps to find a relative higher temperature for each sample, which equals to make student regard the sample as an easier one.

To make the combination method intuitive, we describe the process in Algorithm 1. We conduct first DTD and then KA to modify the distillation training process. (The combination is actually flexible. We can firstly correct teacher logits and then compute sample-wise temperature.) Additionally, to avoid softening repeatedly, LSR replacement should go after distilling softmax, while PS does not mind the order. Thus we place KA after softening for expression consistency.

\begin{algorithm}[!h]
	\caption{DTD-KA Loss Computation}%算法标题
	{\bf Input:}
	teacher logits $\bm{t}$;
	student logits $\bm{v}$;
	ground truth $\bm{y}$;
	{\bf Output:}
	training loss $\mathcal{L}_{total}$;
	\begin{algorithmic}[1]%一行一个标行号
		\STATE get weights $\bm{\omega}_x$ via FLSW or CWSM according to $\bm{t}$;
		\STATE for each sample $x$ compute $\tau_x$ with Eq. (6);
		\STATE calculate softened logits $\bm{q}_{\tau_x}$ and $\bm{q}_{\tau_x}$ using KD softmax function of temperature $\tau_x$;
		\STATE get the sample $i$ whose $\bm{q}_{\tau_i}$ does not match $\bm{y}$;
		\STATE replace $\bm{q}_{\tau_i}$ with Eq. (4) or adjust $\bm{q}_{\tau_i}$ with PS, getting $A(\bm{q}_{\tau_i})$;
		\STATE compute $\mathcal{L}_{total}$ with Eq. (10);
		\STATE return $\mathcal{L}_{total}$;
	\end{algorithmic}
\end{algorithm}

\begin{figure*}[t]
	\begin{center}
		\includegraphics[width=17cm]{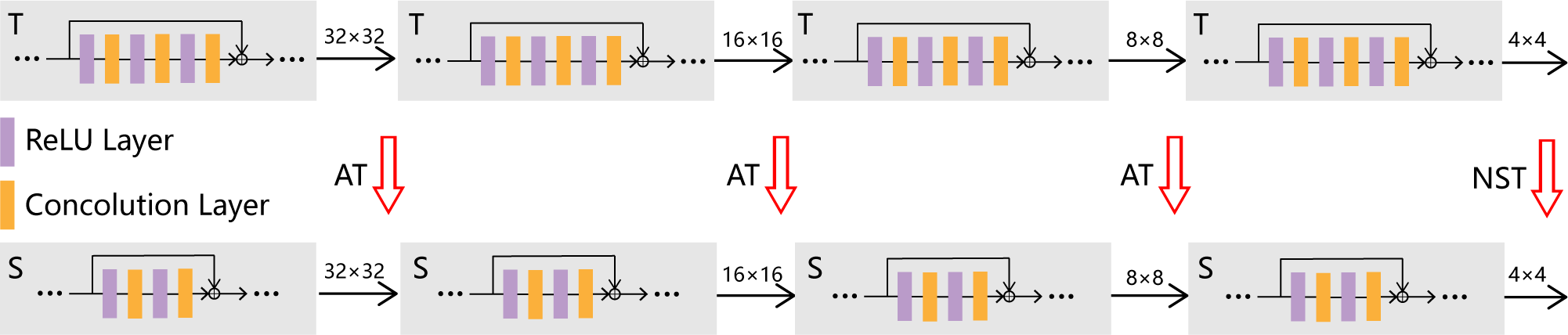}
	\end{center}
	\caption{Transfer pairs of AT and NST for CIFAR-100 experiments. The upper teacher model and lower student model can be clearly divided into 4 stages. Feature maps' resolution keeps unchanged inside each stage. AT loss is computed using the first 3 pairs, and NST loss is computed using the last pair.}
	\label{fig:cifar}
\end{figure*}

\section{Experiments}

In the following section we firstly conduct experiments on CIFAR-100 classification dataset, then validate the method on CINIC-10 \cite{darlow2018cinic} and Tiny ImageNet \cite{le2015tiny}, which are constructed based on ImageNet \cite{russakovsky2015imagenet} but much lighter, for ImageNet training requires vast resources. (Distillation training actually infers twice, once in teacher and once in student in one iteration, thus more time and memory are required.) For each dataset we compare our methods with the conventional KD and some state-of-the-art KD-based works, which are introduced below. Considering different papers adopt various teacher-student pairs, we replicate these methods in the same experimental conditions. 

{\bf Standard KD \cite{hinton2015distilling}.} We adopt Eq. (2) to carry out KD training, where $\alpha$ is set to 0.7. And we implement the cross entropy between two distributions with KL divergence as a convenience.

{\bf AT \cite{zagoruyko2016paying}.} AT introduces attention mechanism to KD and achieves an obvious performance improvement. As a comparison, we transfer activation-based attention information combined with KD, whose loss is
\begin{equation}\label{(11)}
\mathcal{L}_{AT}=\lambda_{AT}\sum_j||\bm{A}_S^j-\bm{A}_T^j||_2+\mathcal{L}_{KD},
\end{equation}
where $\bm{A}_S^j$ and $\bm{A}_T^j$ are the $j$-th pair of L2 normalized attention maps, which respectively come from student model and teacher model.

{\bf NST \cite{huang2017like}.} NST method has recently done solid works and achieved state-of-the-art performance, whose loss is counted using MMD. We adopt the NST+KD combinations as comparison, which performs best among previous works and easy to further modified with our methods. The loss is written in Eq. (12).
\begin{equation}\label{(12)}
\mathcal{L}_{NST}=\lambda_{NST}\mathcal{L}_{MMD^2}(\bm{F}_T, \bm{F}_S)+\mathcal{L}_{KD},
\end{equation}
$\mathcal{L}_{MMD^2}(\bm{F}_T, \bm{F}_S)$ denotes the squared MMD distance between feature maps from certain layers of the teacher model and student model. \cite{huang2017like} proposes several different realizations of $\mathcal{L}_{MMD^2}$, among which the polynomial-kernel-implementation takes most part in experiments. Thus we replicate NST with polynomial kernel for comparison.

{\bf Two Baselines.} We also carry out 2 baseline experiments to help us do comparison. The first one trains the student with standard cross entropy and without teacher, noted as Baseline 1. The another experiment removes the misjudged training samples, noted as Baseline 2, which helps to figure out if these samples influence the performance and genetic errors.  

To avoid performance jittering, we keep initializing seed constant for students in the same structure, and carry out each experiment for several times. 

\begin{table*}[t]
	\centering
	\caption{Results of CIFAR-100 experiments}
	\begin{tabular}{ccccc} % 控制表格的格式
		\toprule[1pt]
		{\bf Method} & {\bf Model} & {\bf with DTD-KA} & {\bf Validation Accuracy} & {\bf Genetic Errors/Total Errors} \\
		\midrule
		
		Teacher & Preact-ResNet-50 & --- & 77.01 & --- \\
		Baseline 1 & Preact-ResNet-18 & --- & 73.76 & ---\\
		Baseline 2 & Preact-ResNet-18 & --- & 76.03 & 1007/2397 = 42.01\%\\
		\multirow{2}*{KD \cite{hinton2015distilling}} & Preact-ResNet-18 & $\times$ & 75.87 & 1022/2413 = 42.35\% \\
		& Preact-ResNet-18 & \checkmark & 77.24 & 950/2276 = 41.74\% \\
		\multirow{2}*{AT+KD \cite{zagoruyko2016paying}} & Preact-ResNet-18 & $\times$ & 76.26 & 1054/2374 = 44.40\% \\
		& Preact-ResNet-18 & \checkmark & 78.06 & 959/2194 = 43.65\% \\
		\multirow{2}*{NST+KD \cite{huang2017like}} & Preact-ResNet-18 & $\times$ & 78.12 & 1087/2188 = 49.68\% \\
		& Preact-ResNet-18 & \checkmark & 78.38 & 1020/2162 = 47.19\% \\
		\bottomrule[1pt]
	\end{tabular}
	\label{tbl:table1}
\end{table*}

\subsection{CIFAR-100}
In CIFAR-100 experiments, we use the official divide of training data and test data, which respectively consist of 50,000 images and 10,000 images, both at a resolution of $32\times32$. Random cropping, random horizontal flipping and random rotation are carried out for data argumentation. We design teacher and student network based on ResNet with pre-activation residual units \cite{he2016identity}, specifically Preact-ResNet50 for teacher and Preact-ResNet18 for student. The student model is optimized for 200 epoches using SGD with mini-batch size of 128, whose momentum is 0.9 and weight decay parameter is $5\times10^{-4}$. The start learning rate is set to 0.01 for KD and the proposed method, 0.1 for AT, NST and combinations. For all experiments the learning rate falls with coefficient 0.1 at 60, 120, 160 epoch.

For LSR, a smoothing factor $\epsilon$ of 0.985 is suitable after experimental attempts. And LSR acts only on misjudged samples, which differs from that in \cite{szegedy2016rethinking}. For DTD, $\tau_0$ and $\beta$ of Eq. (6) are respectively set to 10 and 40. And there are some extra operations. It is observed that $\tau_x$ might get very tiny, even turns to negative in some extreme cases. Thus we bound $\tau_x$ with a threshold of 3. We also cancel the batch average when computing loss. Because we find that average operation could counteract sample-wise learning effect, which comes from customizing dynamic loss computation.

To implement AT and NST, we orderly pair the respective building layers of teacher and student. The residual blocks can be grouped into 4 pairs according to variation of feature map size, just as Fig. 4 shows. The first 3 pairs are chosen to compute $\mathcal{L}_{AT}$, and last pair for $\mathcal{L}_{NST}$. $\lambda_{AT}$ and $\lambda_{NST}$ are set according to suggestions in \cite{szegedy2016rethinking} and \cite{huang2017like}. For experiments of AT+DTD-KA and NST+DTD-KA experiments, we train students using loss of Eq. (13) and Eq. (14), where $\lambda_{AT}^1=\lambda_{AT}^2=1$, $\lambda_{NST}^1=0.1$, and $\lambda_{NST}^2=1$. $\mathcal{L}_{total}$ is expressed as Eq. (10).

\begin{equation}\label{(13)}
\mathcal{L}_{AT}^*=\lambda_{AT}^1\mathcal{L}_{total}+\lambda_{AT}^2\sum_j||\bm{A}_S^j-\bm{A}_T^j||_2.
\end{equation}
\begin{equation}\label{(14)}
\mathcal{L}_{NST}^*=\lambda_{NST}^1\mathcal{L}_{total}+\lambda_{NST}^2\mathcal{L}_{MMD^2}(\bm{F}_T, \bm{F}_S).
\end{equation}

Table 1 shows the CIFAR-100 classification performance. The proposed DTD-KA scheme increases the performance of KD, AT and NST. Simple end-to-end KD with DTD-KA achieve the accuracy of 77.24\%, which is even better than inner-knowledge-focused AT. The NST+DTD-KA combination reaches state-of-the-art. Genetic errors of all the methods are counted. Compared with KD, AT and NST to some extent aggravate the phenomenon because the methods pay more attention to network's intermediate feature, making the student imitate teacher's reference better, no matter whether right or wrong. With bringing strong corrections, our method obviously reduces the genetic errors. And the comparison between baseline 2 and the standard KD shows that misjudged training samples do make nagative effect on performance and genetic errors.

As genetic errors have been reduced, we then investigate performance on uncertain supervision problem. In Fig. 5, we illustrate the accumulation curve of sample along its difference between top 2 student-predicted probabilities. For samples whose difference of top 2 is slight, there exist at leat one disturbance. This denotes that student is lack discrimination or robustness on these samples. It can be observed that methods contain DTD-KA reduce the number of samples who have close top probabilities. It indicates that our method helps the student recognizes more clearly, and means an improvement on robustness from sides.

We also validate the performance of distinct KA and DTD implementations, and results are listed in Table 2. The ablation and crossing combination experiments are designed on KD, AT+KD and NST+KD benchmarks. For each benchmark we firstly add single DTD operation (FLSW/CWSM) or single KA operation (LSR/PS). Then all kinds of DTD-KA combinations are validated. The results show that all implementations indeed make efforts, whether working alone or together.

\begin{figure}[t]
	\begin{center}
		\includegraphics[width=8cm]{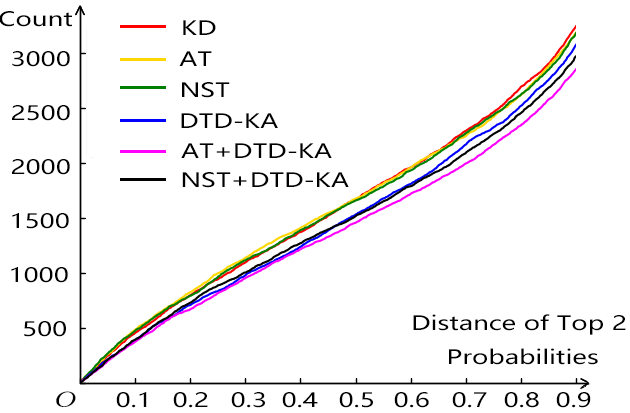}
	\end{center}
	\caption{Statistic of confusing samples. Color lines denote the number of samples whose distance of top 2 predicted probabilities is less than abscissa value}
	\label{fig:fuzzy}
\end{figure}

\begin{table}[t]
	\centering
	\caption{CIFAR-100 ablation and combination experiments. "---" represents no extra operation is conducted.}
	\begin{tabular}{c|c|c|c|c} % 控制表格的格式
		\toprule[1pt]
		{\bf Benchmark} & \diagbox[width=1.5cm]{\bf KA}{\bf DTD} & --- & FLSW & CWSM\\
		\hline
		& --- & 75.87 & 76.57 & 76.63\\
		\multirow{3}*{KD \cite{hinton2015distilling}}& LSR & 76.24 & {\bf 77.24} & 76.88\\
		& PS & 76.05 & 76.72 & 76.79\\
		& --- & 76.26 & 76.98 & 76.80\\
		\multirow{3}*{AT+KD \cite{zagoruyko2016paying}}& LSR & 77.70 & 77.97 & {\bf 78.06}\\
		& PS & 77.75 & 77.89 & 77.78\\
		& --- & 78.12 & 78.33 & 78.24\\
		\multirow{3}*{NST+KD \cite{huang2017like}}& LSR & 78.16 & {\bf 78.38} & 78.30\\
		& PS & 78.31 & 78.36 & 78.37\\
		\bottomrule[1pt]
	\end{tabular}
	\label{tbl:table2}
\end{table}

\begin{figure*}[t]
	\begin{center}
		\includegraphics[width=17cm]{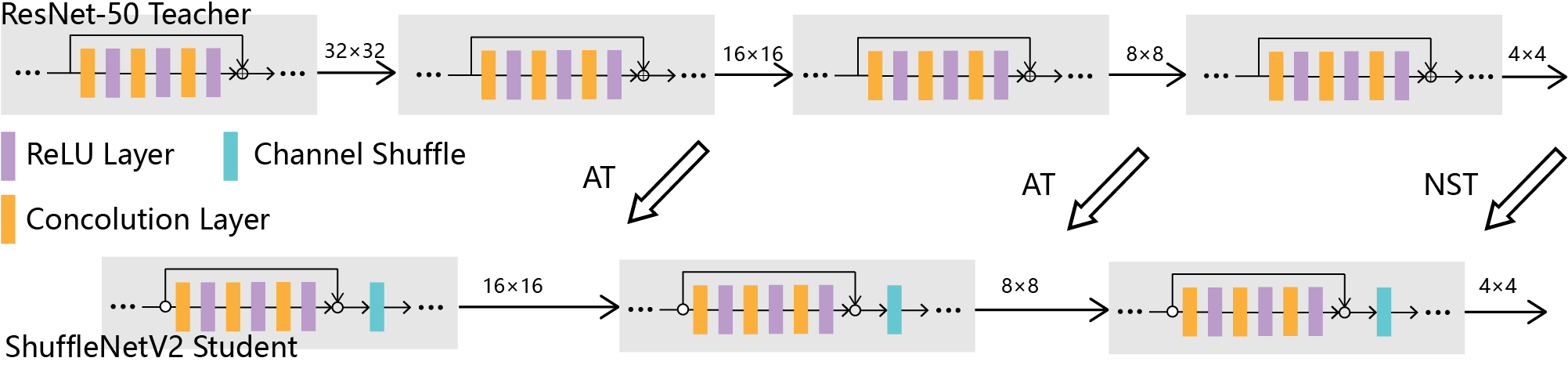}
	\end{center}
	\caption{Transfer pairs of AT and NST for Tiny ImageNet experiments. The ResNet teacher's last three stages are respectively paired with the ShuffleNet student's three stages. AT loss is computed using the first 2 pairs, and NST loss is computed using the last pair.}
	\label{fig:tiny}
\end{figure*}

\begin{table*}[t]
	\centering
	\caption{Results of CINIC-10 experiments.}
	\begin{tabular}{ccccc} % 控制表格的格式
		\toprule[1pt]
		{\bf Method} & {\bf Model} & {\bf with DTD-KA} & {\bf Validation Accuracy} & {\bf Genetic Errors/Total Errors} \\
		\midrule
		Teacher & ResNet-50 & --- & 87.48 & --- \\
		Baseline 1 & ResNet-18 & --- & 84.74 & ---\\
		Baseline 2 & ResNet-18 & --- & 86.76 & 5754/11916 = 48.29\%\\
		\multirow{2}*{KD \cite{hinton2015distilling}} & ResNet-18 & $\times$ & 86.52 & 5891/12130 = 48.57\% \\
		& ResNet-18 & \checkmark & 87.02 & 5527/11678 = 47.33\% \\
		\multirow{2}*{AT+KD \cite{zagoruyko2016paying}} & ResNet-18 & $\times$ & 87.03 & 6094/11669 = 52.22\% \\
		& ResNet-18 & \checkmark & 87.33 & 5703/11395 =50.05\% \\
		\multirow{2}*{NST+KD \cite{huang2017like}} & ResNet-18 & $\times$ & 86.97 & 6259/11730 = 53.36\% \\
		& ResNet-18 & \checkmark & 87.17 & 5983/11549 = 51.81\% \\
		\bottomrule[1pt]
	\end{tabular}
	\label{tbl:table3}
\end{table*}

\begin{table*}[t]
	\centering
	\caption{Results of Tiny ImageNet experiments.}
	\begin{tabular}{ccccc} % 控制表格的格式
		\toprule[1pt]
		{\bf Method} & {\bf Model} & {\bf with DTD-KA} & {\bf Validation Accuracy} & {\bf Genetic Errors/Total Errors} \\
		\midrule
		Teacher & ResNet-50 & --- & 61.90 & --- \\
		Baseline 1 & ShuffleNetV2-1.0 & --- & 60.05 & ---\\
		Baseline 2 & ShuffleNetV2-1.0 & --- & 61.38 & 1191/3862 = 30.84\%\\
		\multirow{2}*{KD \cite{hinton2015distilling}} & ShuffleNetV2-1.0 & $\times$ & 61.31 & 1348/3910 = 34.48\% \\
		& ShuffleNetV2-1.0 & \checkmark & 61.77 & 1144/3823 = 29.92\% \\
		\multirow{2}*{AT+KD \cite{zagoruyko2016paying}} & ShuffleNetV2-1.0 & $\times$ & 61.27 & 1375/3873 = 35.50\% \\
		& ShuffleNetV2-1.0 & \checkmark & 61.47 & 1261/3853 = 32.73\% \\
		\multirow{2}*{NST+KD \cite{huang2017like}} & ShuffleNetV2-1.0 & $\times$ & 60.83 & 1399/3917 = 35.72\% \\
		& ShuffleNetV2-1.0 & \checkmark & 61.19 & 1283/3881 = 33.06\% \\
		\bottomrule[1pt]
	\end{tabular}
	\label{tbl:table4}
\end{table*}

\subsection{CINIC-10}

CINIC-10 extends CIFAR-10 dataset with downsampled ImageNet images, which consists of 90,000 training images, 90,000 validation images and 90,000 test images, all at a resolution of $32\times32$. We adopt the same data preprocessing as those of CIFAR-100 experiments. The teacher model and student model here are constructed using ResNet \cite{he2016deep} of 50 layers and 18 layers. The training lasts for 200 epoches and the learning rate falls along cosine annealing \cite{loshchilov2016sgdr}. We also set the SGD optimizer with batch size 128, momentum 0.9 and weight decay $5\times10^{-4}$. Parameters in computing loss keep aligned with CIFAR-100 experiments, except for $\lambda_{AT}^1=0.01$, $\lambda_{AT}^2=1$, $\lambda_{NST}^1=0.01$, and $\lambda_{NST}^2=1$. The above settings help to numerically balance the contributions of $\mathcal{L}_{total}$, $\mathcal{L}_{AT}$ and $\mathcal{L}_{NST}$, since the KL divergence goes enormous when the number of classes is small (CINIC-10 has 10 classes).

In Table 3 we compare the CINIC-10 performance among KD, AT+KD, NST+KD and the proposed methods. As a simple end-to-end distillation method, DTD-KA shows competitive performance compared with inner-feature-based methods AT and NST. And obviously, all previous works get increase when served by our methods. Furthermore, genetic errors also get reduced.

While it can be observed that all the improvement methods seem not to increase KD as much as that in CIFAR-100 experiments. Besides data issues, we guess it is because activated feature maps might be weaker than nonactivated ones, which are transferred in Section 4.1. According to \cite{sandler2018mobilenetv2}, feature maps after activation (ReLU) might lose some information, especially in low channel case. Thus we assume that feature maps before rather than after activation might be more suitable for knowledge transfer, which certainly deserves more future discussions.

\subsection{Tiny ImageNet}

Tiny ImageNet selects 120,000 images from ImageNet and downsamples them to $64\times64$. These images come from 200 categories, each of which has 500 training images, 50 validation images, and 50 test images.

On Tiny ImageNet we validate performance of transferring knowledge with heterogeneous teacher and student, which stumps most KD-based methods. A ResNet-50 teacher model and a student model implemented in ShuffleNet V2-1.0 \cite{ma2018shufflenet} are adopted as subjects. We pair the last three ResNet building stages and the three ShuffleNet stages, whose output feature maps match in resolution, thus no additional mapping operation is needed. The process is illustrated in Figure 6. Student is trained for 200 epoches with learning rate starting from 0.01 and decaying with cosine annealing. And SGD of batch size 128, momentum 0.9 and weight decay $5\times10^{-4}$ is adopted. $\lambda_{AT}^1$, $\lambda_{AT}^2$, $\lambda_{NST}^1$ and $\lambda_{NST}^2$ are respectively set to 0.08, 1, 0.08 and 10.  A relatively small weight for $\mathcal{L}_{total}$ can help student to find correct convergence direction. And DTD-KA loss is computed also with the sum rather than the average for one batch.

Table 4 summarizes the comparison results. The proposed DTD-KA gets highest accuracy and lowest genetic errors. We can observe regressions of AT+KD and NST+KD. As predicted above, the inner-feature-based methods perform weaker than end-to-end methods. Nevertheless, DTD-KA still improves them and reduced genetic errors.

Actually there always lies a gap between student and teacher. For AT, NST and other inner-feature-based approaches, the raw intermediate information from a heterogeneous teacher may not be reliable enough for target student. While end-to-end methods are more robust, since these methods do not impose strict constraints on inference process. Our results prove that better and stronger soft targets may help in such kinds of circumstances.

\subsection{Discussion}
Cubersome teacher networks generally perform well on training set, which indicates there are actually a small number of misjudged samples. We find from experiment results that these samples with proper correction can lead to an observable increase on performance. The decrease of genetic errors on Tniy-ImageNet is much more than that on CIFAR-100 and CINIC-10, which might come from a higher proportion of misjudged samples---Teacher network doesn't behave as well on Tiny-Imagenet as it does on the other two datasets.

In KA, we fix the incorrect predictions according to the ground truth label. And this needs premises: labeled data is necessary and the labels are assumed to be totally correct, which are hard to meet in applications. Therefore, how label regularization without dependence on ground truth behave in KA deserves future investigation. (KA actually carries out combinations of label regularization methods, KD and some other.) As for DTD, it deals with model prediction on samples, so it works on unlabeled or imperfectly labeled data.

\section{Conclusion}

Supervision plays a significant role in knowledge distillation (or transfer) works. We propose Knowledge Adjustment (KA) and Dynamic Temperature Distillation (DTD), trying to answer the challenge of bad supervision in existing KD and KD-based methods. We divide the problem into incorrect supervision and uncertain supervision, for both phenomenons we give our comprehension and propose our corresponding solutions. KA corrects teacher's wrong predictions according to ground truth. DTD fixes teacher's uncertain predictions with dynamic temperature. Validation experiments on three different datasets prove that the proposed methods can increase accuracy. Statistical results further indicate the proposed methods indeed reduce genetic errors and improve student on discrimination. In addition, combination experiments show that our methods can easily attach to other KD-based methods. We believe that our methods can be applied in many knowledge distillation frameworks, and sample-based idea of this paper might lead to further progress.

\ifCLASSOPTIONcaptionsoff
  \newpage
\fi



%\bibliographystyle{IEEEtran}
%\begin{thebibliography}{1}
%\bibitem{s}sss
%\end{thebibliography}
\bibliographystyle{IEEEtran}
\bibliography{IEEEabrv,mybib}
\end{document}